\documentclass[conference]{IEEEtran}
\IEEEoverridecommandlockouts
\usepackage{cite}
\usepackage{amsmath,amssymb,amsfonts}
\usepackage{algorithmic}
\usepackage{graphicx}
\usepackage{tabularx}
\usepackage{array}
\usepackage{textcomp}
\usepackage{xcolor}
\usepackage{subcaption}
\def\BibTeX{{\rm B\kern-.05em{\sc i\kern-.025em b}\kern-.08em
    T\kern-.1667em\lower.7ex\hbox{E}\kern-.125emX}}
    
\newcolumntype{C}[1]{>{\centering\let\newline\\\arraybackslash\hspace{0pt}}m{#1}}
\begin{document}

\title{Mobile Robot Planner with Low-cost Cameras Using Deep Reinforcement Learning\\
% {\footnotesize \textsuperscript{*}Note: Sub-titles are not captured in Xplore and should not be used}
% \thanks{Identify applicable funding agency here. If none, delete this.}
}

% \author{\IEEEauthorblockN{Quang-Minh Tran\textsuperscript{1}}
% \IEEEauthorblockA{\textit{Faculty of Information Technology} \\
% \textit{dept. Computer Vision and Cognitive Cybernetics}\\
% \textit{VNUHCMC-University of Science}\\
% Ho Chi Minh City, Vietnam \\
% 1612374@student.hcmus.edu.vn}
% \and
% \IEEEauthorblockN{Ngoc Quoc Ly\textsuperscript{2}}
% \IEEEauthorblockA{\textit{Faculty of Information Technology} \\
% \textit{dept. Computer Vision and Cognitive Cybernetics}\\
% \textit{VNUHCMC-University of Science}\\
% Ho Chi Minh City, Vietnam \\
% lqngoc@fit.hcmus.edu.vn}
% }
\author{
\IEEEauthorblockN{
Minh Q. Tran\IEEEauthorrefmark{1}, and
Ngoc Q. Ly\IEEEauthorrefmark{2}}
\IEEEauthorblockA{
\textit{
\IEEEauthorrefmark{1}\IEEEauthorrefmark{2}Faculty of Information Technology, 
\IEEEauthorrefmark{2}Computer Vision \& Cognitive Cybernetics Dept.} \\
\textit{VNUHCM-University of Science}\\
Ho Chi Minh City, Vietnam \\
\IEEEauthorrefmark{1}1612374@student.hcmus.edu.vn,
\IEEEauthorrefmark{2}lqngoc@fit.hcmus.edu.vn}
}

% \and
% \IEEEauthorblockN{3\textsuperscript{rd} Given Name Surname}
% \IEEEauthorblockA{\textit{dept. name of organization (of Aff.)} \\
% \textit{name of organization (of Aff.)}\\
% City, Country \\
% email address or ORCID}
% \and
% \IEEEauthorblockN{4\textsuperscript{th} Given Name Surname}
% \IEEEauthorblockA{\textit{dept. name of organization (of Aff.)} \\
% \textit{name of organization (of Aff.)}\\
% City, Country \\
% email address or ORCID}
% \and
% \IEEEauthorblockN{5\textsuperscript{th} Given Name Surname}
% \IEEEauthorblockA{\textit{dept. name of organization (of Aff.)} \\
% \textit{name of organization (of Aff.)}\\
% City, Country \\
% email address or ORCID}
% \and
% \IEEEauthorblockN{6\textsuperscript{th} Given Name Surname}
% \IEEEauthorblockA{\textit{dept. name of organization (of Aff.)} \\
% \textit{name of organization (of Aff.)}\\
% City, Country \\
% email address or ORCID}

\maketitle

\begin{abstract}
This study develops a robot mobility policy based on deep reinforcement learning. Since traditional methods of conventional robotic navigation depend on accurate map reproduction as well as require high-end sensors, learning-based methods are positive trends, especially deep reinforcement learning. The problem is modeled in the form of a Markov Decision Process (MDP) with the agent being a mobile robot. Its state of view is obtained by the input sensors such as laser findings or cameras and the purpose is navigating to the goal without any collision. There have been many deep learning methods that solve this problem. However, in order to bring robots to market, low-cost mass production is also an issue that needs to be addressed. Therefore, this work attempts to construct a pseudo laser findings system based on direct depth matrix prediction from a single camera image while still retaining stable performances. Experiment results show that they are directly comparable with others using high-priced sensors.
\end{abstract}


\section{Introduction}
There have been many studies on the robot navigation problem, spread across many different methods with certain achievements. The first of all is the method based on the map \cite{mapbased1}, \cite{mapbased2}, which requires a given map of the environment, thereby making decisions to navigate the behavior of the environment. However, due to the fact that not all environments have available map description, a lot of research approaching map reconstruction \cite{mapreconstruct1}, \cite{mapreconstruct2}, \cite{mapreconstruct3}, \cite{ humanguide1} have emerged. However, they require a lot of precision and cost, to ensure the route-finding algorithms and robot movements can be most effectively applied.\newline

Accordingly, mapless-based algorithms are also very popular with modules namely mapping, localization, planning, and locomotion. Simultaneous Localization And Mapping (SLAM) is a typical problem for solving the first two modules above, thereby establishing a path for motion. The strength of this approach is that the robot reproduces the map by itself using the power of high-end sensors. From there, it will not be much difficult for the robot to move according to a route-finding algorithm in the environment created. There are several challenges to this approach, including the time-consuming building and updating maps, as well as the hardware requirements to achieve high precision.\newline

In order to reduce the cost of hardware necessities, research on robot navigation based solely on visual information from cameras is more intriguing in research and also offers a lot of promises. Methods focusing on avoiding obstacles with given input images, such as \cite{potential_fields}, \cite {sonar}, \cite{visual_mem}, \cite{vision_3d} are reasonably potential. In parallel, to lessen the refinement requirements and dependence on specialized knowledge, some methods combine the modules into a ``black box'', mapping directly from the observed state to the robot movement based on learning algorithms. Indeed, deep learning has the potential to solve the approximation problem as well, \cite{Chen_2015_ICCV} applied it to extract semantic information from the camera image to decide on actions like turn left or right. However, since supervised learning gives a specific action ground truth that corresponds to each state observed by a robot, this method has good results but the surrounding environment should be tightly constrained and not highly generalized. Meanwhile, reinforcement learning posses the more general problem modeling by rewarding the robot with the action it performs according to the observation. Therefore, the action set can be more diverse, and the problem is more generalized and practically applicable.\newline

In this work, the authors choose to solve the problem by applying deep reinforcement learning based on the method of this study \cite{tai2017virtual}. On that basis, we leverage one of the state-of-the-art algorithms, being Soft Actor-Critic (SAC) \cite{haarnoja2018soft}, to solve the MDP with continuous action space domain in the training phase. This training strategy improves directly from the Deep Deterministic Policy Gradient (DDPG) \cite{lillicrap2015continuous} algorithm which \cite{tai2017virtual} used in their research. SAC is also applied by \cite{chaffre2020sim} to the same problem. Next, also the main contribution of our paper is geared towards using images from a monocular camera, instead of using costly input sensors. The main purpose of this replacement is to save costs on hardware since the prices of laser-emitting devices or cutting-edge cameras are often much higher. Indeed, there have been studies of \cite{xie2017towards} and \cite{ma2019using} using CNN to derive the depth map from the RGB image and taking it as the input state of the robot. However, instead of using the whole depth map, we lessen the state vector size by sampling only the $n$ depth values from a specific row of it. The depth prediction phrase is inherited from \cite{godard2019digging}, a network architecture pre-trained on KITTI \cite {geiger2013vision} dataset. From there, we applied transfer learning so that the results attained could be comparable with the method using the laser findings.\newline

In particular, our contributions can be summarized as follows: (i) We analyse the gains of Soft Actor-Critic to solve the MDP with continuous action space domain; (ii) We propose using images from a monocular camera, instead of using costly input sensors to save costs on hardware; and (iii) We apply transfer learning to make sure that the results attained is comparable with the method using laser findings. In Section II, the paper reviews the previous work in deep reinforcement learning as well as its application in robot navigation. Section III describes the baseline and our proposed methods. Section IV shows the experimental results. Conclusion and future works will be presented in Section V. 

\section{Related works}

\subsection{Deep reinforcement learning}
Deep learning has grown strongly in the past decade with the breakthrough of AlexNet \cite{NIPS2012_4824} on the Image Net \cite{imagenet_cvpr09} dataset, leading to a series of new deep learning network architectures \cite{szegedy2014going}, \cite{simonyan2014deep}, \cite{he2015deep}. Besides, the introduction of new optimization techniques \cite{ruder2016overview} together with transfer learning methods \cite{tan2018survey}, and the development of computer hardware supporting parallel computation.\newline

Follow the achievements of deep learning, when it well solves the functional approximation problem, deep reinforcement learning also emerged with excellent works. Starting with  Deep Q Learning (DQN) \cite{mnih2013playing}, the algorithm uses neural networks to approximate the function Q, combined with replay memory, storing state transitions at different times. This algorithm is an upgraded version of Q-learning, in which the target policy differs from the behavior policy used to collect state transitions in the trajectory. The main advantage of the off-policy or Q-learning approach comes from the fact that it does not need the entire trajectory, from the start state to the end, and can reuse past state transitions from the replay memory, which provides efficiency in sampling. Another advantage is that because of the off-policy property, it helps the algorithm maintain exploration to deal with the exploration-exploitation trade-off in reinforcement learning.\newline

However, the weakness of the off-policy techniques is the lack of stability and the difficulty of convergence, whereas the on-policy methods directly update target policy via maximizing the expected return so making the convergence more stable and reliable. Trust Region Policy Optimization algorithm (TRPO) \cite{schulman2015trust} came into being, based on the actor-critic model with the constraint of policy updates measured by the Kullback Leibler divergence, allowing for larger training step and accelerating the convergence process. By improving TRPO, Proximal Policy Optimization (PPO) \cite{schulman2017proximal} points out the difficulties in implementing TRPO then uses a surrogate loss function, which supports convergence acceleration. The algorithms above, in contrast to off-policy ones,  are stable but lack efficiency in sampling when the entire trajectory is required.\newline

In order to balance the strengths and weaknesses of the two approaches mentioned above, there have been techniques, parallel approximate the target policy and also the action-value function based on the Bellman equation. These algorithms often solve the MDP with continuous action domains and Deep Deterministic Policy Gradient (DDPG) \cite{lillicrap2015continuous}, Rainbow Q-learning \cite {hessel2017rainbow}, T3D \cite{fujimoto2018addressing}, SAC \cite {haarnoja2018soft} are such algorithms. Therefore, for the problem of this study, we take advantage of these algorithms.
\subsection{Deep reinforcement learning in Robot Navigation}
Reinforcement learning is widely applied to robots in general. \cite{kohl2004policy} uses a policy optimization approach to control the movement of a quadruped robot, while, with a similar method, \cite{peters2008reinforcement} discusses primitive locomotion. \cite{michels2005high} applies the model-based algorithm for obstacle avoidance problems with a single camera image or \cite{kim2004autonomous} for helicopter automation.\newline

In terms of mapless navigation, deep reinforcement learning is quite prevalent. \cite {zhu2017target} used the actor-critic model combined with target image information as an implicit input that makes the method generalized with the same type of environment without the need for retraining from scratch. \cite{mirowski2016learning} applied a policy optimization method with depth image estimation in parallel with trained supervised learning in an extremely complex and comprehensive simulation environment. \cite{ma2019using}, \cite{tp2018learning} with PPO also has similar approaches. \cite{tai2017virtual} uses the DDPG algorithm to solve the problem of navigating to the destination and avoiding obstacles for mobile robots with laser findingss and can be integrated into the real environment. Also solves the problem of mobile robots with a monocular camera, \cite{xie2017towards} leverage image depth information predicted from a convolutional network, incorporate with Dueling DQN, trained on many simulation environments from simple to complex and then put into a real environment.
\section{Method}
This paper implements \cite{tai2017virtual} as the baseline with the same problem statement. Base on this one, our proposal aims to replace the robot's observation.
\subsection{Problem definition}
The objective of the problem is to find the optimal policy, which maps directly from the state of the robot to the action it will perform at each time step $t$.
\begin{equation}
    a_t = \pi(s_t)
\end{equation}
Each state $s_t$ is comprised of three factors, namely the observation $o_t$ obtained from the input sensor, values in polar coordinates $p_t$ represent the position of the target, and the action $a_{t-1}$ performed previously.
\begin{equation}
    s_t = [o_t, p_t, a_{t-1}]
\end{equation}

So as to reinforce the action, at each time step, a reward $r$ is defined, relying on the state attained by that behavior as follows.
\begin{equation}
    r = R(s_t, a_t, s_{t+1})= 
\begin{cases}
    r_{arrive},& \text{if } d_{t+1} \leq c_d\\
    r_{collision}, & \text{if } is\_collided\\
    c_r(d_{t} - d_{t+1} ), & \text{otherwise}\\
\end{cases}
\label{reward_eq}
\end{equation}
$r_{arrive}$ is the reward achieved when the distance from the robot to the target $d_{t+1}$ is less than the given threshold $c_d$. By contrast, $r_{collision}$ is the negative feedback to be received when the $is\_collided$ is true returned by the bumper. If neither of the above cases is concerned, the reward will be the subtraction of the distance to the destination point earlier and the one later multiplied by a hyper-parameter to be fine tuned $c_r$. This reward will be is negative if that gap has not been shortened over time.
\subsection{DDPG and SAC}
We recall the action-value function $Q$. The purpose of this function is to evaluate whether a policy $\mu$ is good or not. Indeed, $Q(s,a)$ is the expected return of trajectories starting with state $s$ and action $a$, then other states and actions are sampled by the policy $\mu$.
\begin{equation}
    \tau = (s, a, r_1, s_1, \mu(s_1), ..., s_{T-1}, \mu(s_{T-1}), r_T, s_T),
\end{equation}
\begin{equation}
    Q_{\mu_\theta}(s,a) = \underset{\tau \sim \mu}{\mathbb{E} } [G(\tau)|s_0 = s, a_0 = a]
\end{equation}

The policy giving the maximum $Q$ will be the optimal one, denoted $\mu^*$. This $Q$ is then called the optimal action value function, briefly denoted $Q^*$. Furthermore, this function $Q^{*}$ always satisfies an equation, called the Bellman equation, as shown below.
\begin{equation}
    Q^{*}(s,a) \approx r + \max_{a'}Q^{*}(s', a')
    \label{bellmaneq}
\end{equation}

\begin{center}
    \begin{figure}[htbp]
    \begin{center}
     \includegraphics[scale=.48]{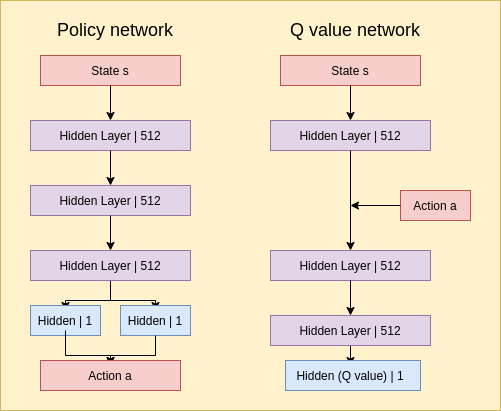}
    \end{center}
    \caption{Parameterized policy and Q value neural networks}
    \label{policy_Q_nets}
    \end{figure}
\end{center}

As well as leveraging \ref{bellmaneq} like DQN did to approximate the function $Q^*$ by parameterizing it with $\phi$, DDPG also parameterized the policy with the weights $\theta$. Indeed, $\theta$ is updated through the optimization problem \ref{theta_loss_func}. In addition, $\theta$ and $\phi$ is designed as neural networks described in the figure \ref{policy_Q_nets}.
\begin{equation}
    \theta^{*} = \underset{\theta}{\arg\max} \underset{s \sim \mathcal{D}}{\mathbb{E}}\big[\mathit{Q}_{\phi}(s,\mu_\theta(s))\big]
    \label{theta_loss_func}
\end{equation}

SAC, an improvement on DDPG, tackle two common issues, that off-policy approaches have to deal with. First, SAC has better control over the exploration-exploitation trade-off. Indeed, the reward at each time step is added a value proportional to the entropy $H$ of $\pi_\theta$ by the factor $\alpha$.
\begin{equation}
    R(s_t, a_t, s_{t+1}) + \alpha H (\pi_\theta(.|s_t))
\end{equation}
Accordingly, maximizing this reward (also maximizing the entropy value) makes the probability of choosing the action from the policy more evenly distributed. This means that the robot will keep exploring instead of just focusing on exploiting the reward at the moment. The value $\alpha$ can be adjusted during training to increase or decrease the level of exploration. Therefore, Bellman equation also has some modification corresponding to the new rewards, specifically as follows
\begin{equation}
\begin{split}
    Q(s,a) \approx r + \gamma (Q(s',\tilde{a}') + \alpha H\pi(.|s'))\\
    \approx r + \gamma (Q(s',\tilde{a}') - \alpha \log \pi(\tilde{a}'|s'))
\end{split}
\end{equation}
where the action at each point is sampled from the random policy $\pi_\theta$, which is a Gaussian distribution with the expected value $\mu$ and the variance $\sigma$ is parameterized below the form of deterministic functions with $\theta$. $\mathcal{N} (0,I)$ is the spherical gaussian distribution.
\begin{equation}
    \tilde{a} = \tanh(\mu_\theta(s) + \sigma_\theta(s) \odot \xi ), ~~\xi \sim \mathcal{N} (0,I)
\end{equation}

The second improvement of the SAC is that it helps to stabilize the output of the action-value function. In DDPG, the learned function $Q$ tends to estimate the return to be greater than the expected one. To tackle this issue, SAC implements two action-value functions $Q_{\phi_{1}}, Q_{\phi_{2}}$, and uses the smaller one as the predicted return. Due to the above two enhancements, we leverage SAC as the main learning method to solve the problem.
\subsection{Proposal}
We utilize the UNet model in \cite{godard2019digging} to directly predict depth images from monocular images. This UNet provides deep features as well as local information of images. In fact, the encoder uses the ResNet-18 model \cite{he2016deep} comprised of 11 million parameters while the decoder includes upsampling convolution with ELU activation function in each layer and sigmoid function at the last layer to obtain the disparity map $D$. The whole UNet is also pre-trained on the KITTI dataset. \cite{geiger2013vision}.
\begin{center}
    \begin{figure}[htbp]
    \begin{center}
     \includegraphics[scale=.30]{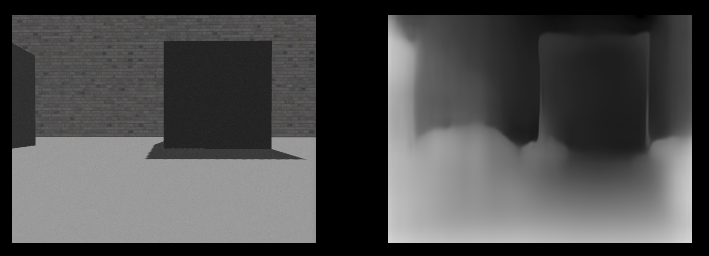}
    \end{center}
    \caption{The image observed by robot monocular camera in Gazebo simulation (left), and disparity map predicted (right)}
    \label{depth_map}
    \end{figure}
\end{center}

The advantage of this model is that while using only the ResNet-18 from which the depth characteristics obtained from the image are less efficient than the other massive models,  it offers a faster speed in inference time. Moreover, the results of the whole model are also competitive compared to other methods due to the fact that it is strongly supported based on the self-supervised learning method, generating labels from existing data. Indeed, this work used the correlation between frames through times to establish loss functions.\newline

Once the disparity map $D$ sized $(w, h)$ is obtained, we sample the inverses of $n$ values $D_{i, j}$, $i \in (0, w), j = h / 2$, from the disparity map and use it as observation.
\begin{equation}
    o_t = \Big[\frac{1}{D_{i_1,j}}, \frac{1}{D_{i_2,j}}, ..., \frac{1}{D_{i_n,j}}\Big]
\end{equation}

The values $\frac{1}{D_ {i, j}}$ are directly proportional to the distance of the camera to the pixel points. This can be considered as building a $n$ pseudo long-distance laser findings system in a much more economical way.

\section{Experimental results}
In this section, we conduct the training and evaluation process in virtual environments, simulated by Gazebo \cite{Aguero-2015-VRC}. The robot chosen was TurtleBot Burger version 3.0. This work also designs training environments as well as evaluation scenarios from simple to complex, thereby evaluating the efficiency of our proposal compared with the baseline.
\subsection{Training}
We designed two training environments in which the robot was placed in a space of about $64m^2$, surrounded by walls. Destinations will appear randomly in this space. Each time when the robot hits an obstacle, or travel over the time limit specified for a trajectory, it is reset back into its original position. Meanwhile, if the robot reaches the goal, a new one will be spawned in a different location. The second training environment is different from the first one with 4 static obstacles located close to the starting position of the robot.\newline
\begin{figure}
\centering
\begin{subfigure}{.25\textwidth}
  \centering
  \includegraphics[width=.9\linewidth]{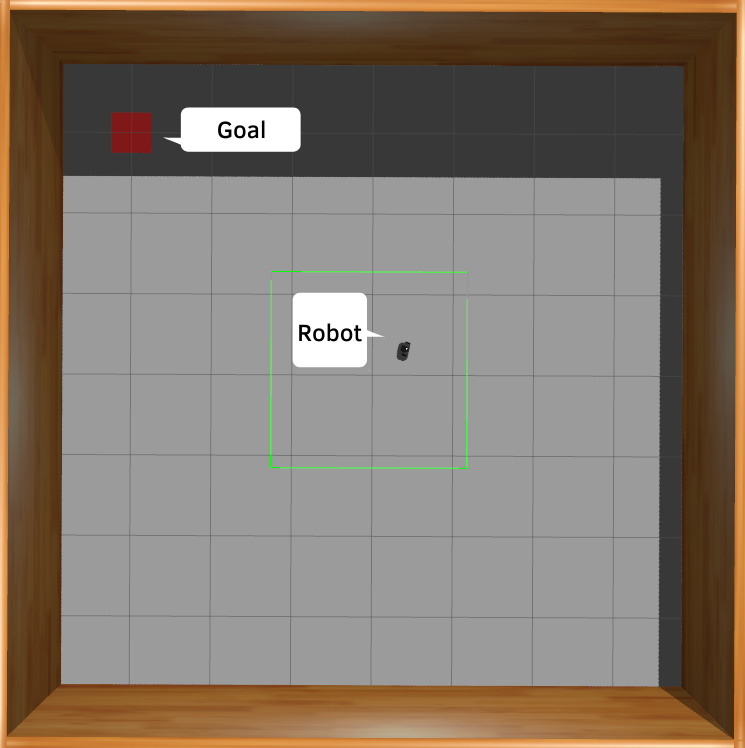}
  \caption{Env-1}
  \label{fig:sub1}
\end{subfigure}%
\begin{subfigure}{.25\textwidth}
  \centering
  \includegraphics[width=.9\linewidth]{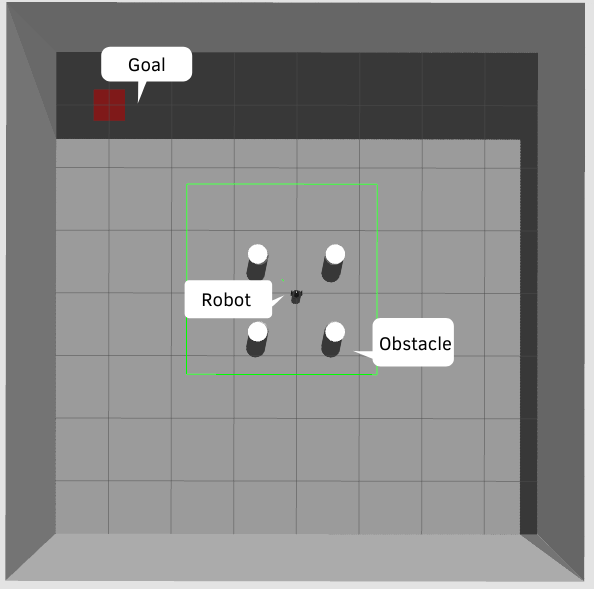}
  \caption{Env-2}
  \label{fig:sub2}
\end{subfigure}
\caption{Training environments simulated in Gazebo}
\label{train_env}
\end{figure}

In terms of the training setup, the values of rewards are: $r_{collision} = -200, r_{arrive} = 150, c_r = 500$. The size of the observation vector from the laser findings or predicted depth map is 10. In addition, the size of the replay buffer $\mathcal{D}$ is 100,000 state transitions $\mathcal{D}_i$ , the batch size for each time transitions sampled from this set is 128. We utilize the Adam \cite{kingma2014adam} algorithm with a learning rate of 0.0001 to solve the optimization problem. Besides, the training time takes about 45 minutes for every 10000 steps performed.

\begin{figure}%
    \centering
    \subfloat[Env-1]{{\includegraphics[width=9cm]{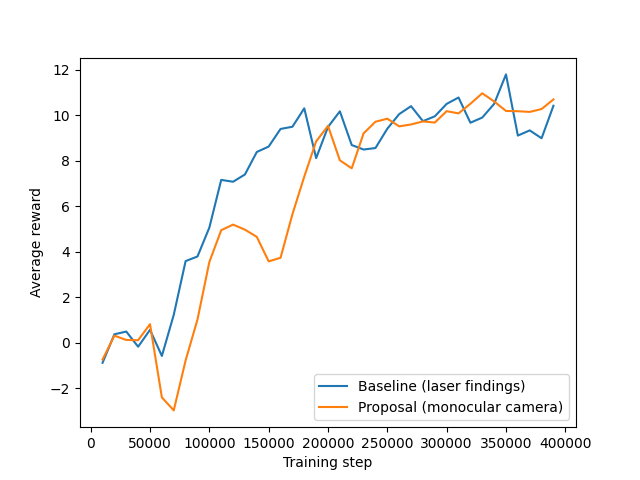} }}%
    \qquad
    \subfloat[Env-2]{{\includegraphics[width=9cm]{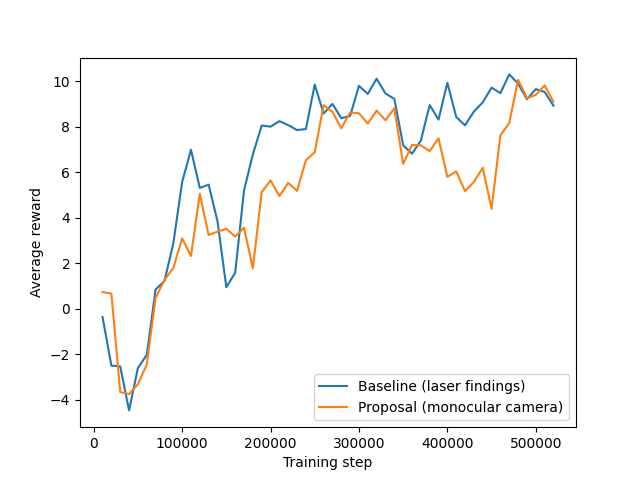} }}%
    \caption{Performance of training phase in two environment. As can be illustrated from the chart, all models have converged to adequate and stable performance. Furthermore, the patterns of both methods are quite similar in most of the time in both environments.}%
    \label{train_figure}%
\end{figure}

\subsection{Evaluation}
In this subsection, we build scenarios on the simulator to simulate the process of evaluation. These are the scenarios that robots have never seen before, from which our methodology can be evaluated.\newline
\begin{figure*}
\centering
\begin{subfigure}{.5\textwidth}
  \centering
  \includegraphics[width=.8\linewidth]{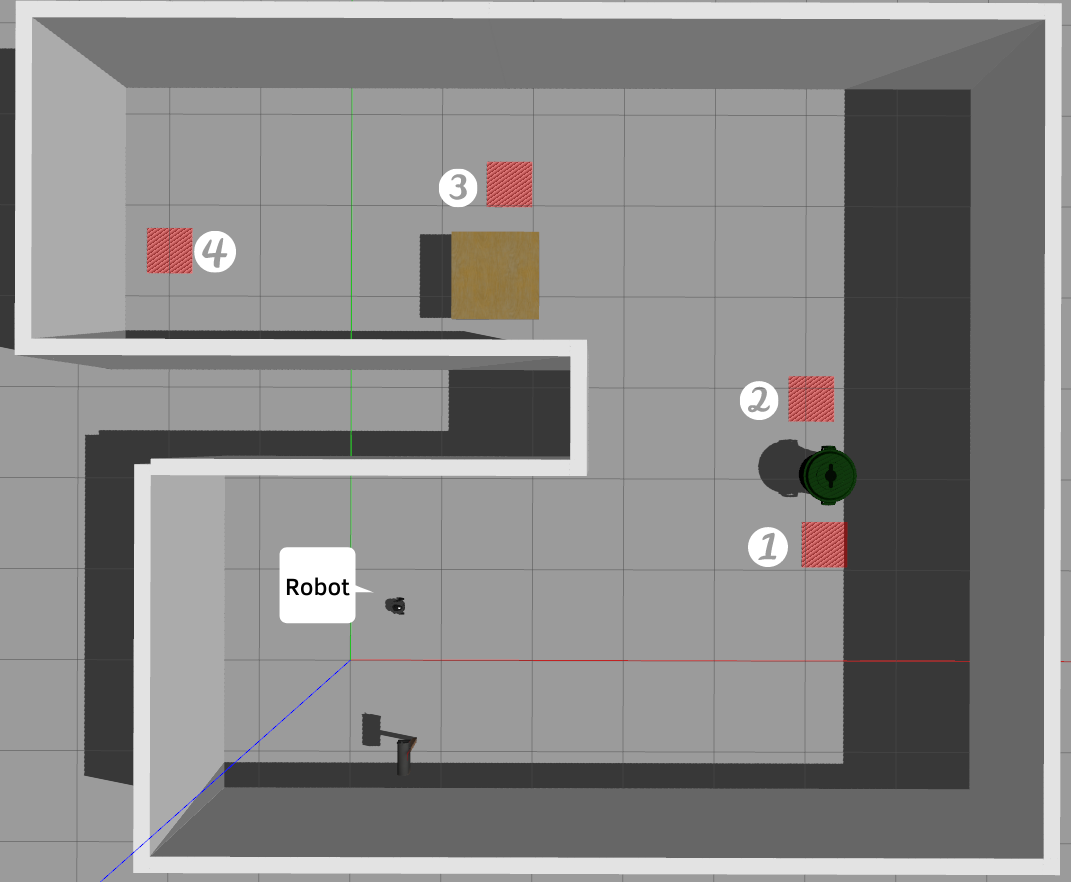}
  \caption{Evaluation scenario 1}
  \label{fig:sub1}
\end{subfigure}%
\begin{subfigure}{.5\textwidth}
  \centering
  \includegraphics[width=.8\linewidth]{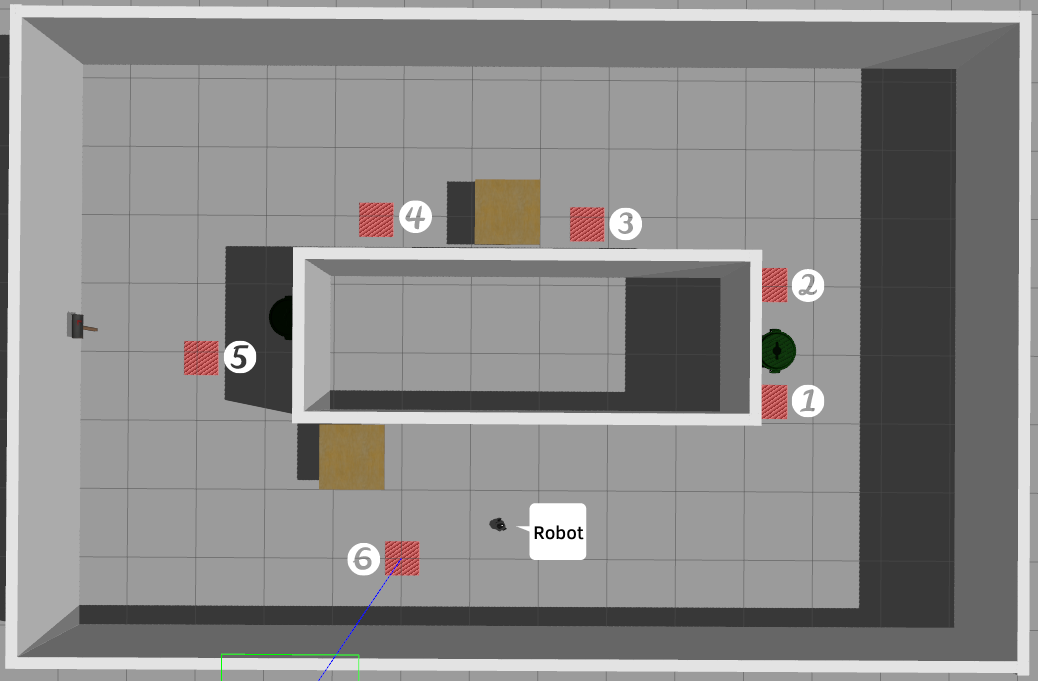}
  \caption{Evaluation scenario 2}
  \label{fig:sub2}
\end{subfigure}
\caption{Evaluation scenarios. The aim is to reach all the red destinations without any collision. Both evaluation scenarios are designed in the form of an office, in which obstacles are placed between the target points to create complexity. The second one is more complicated than the first one with a longer travel distance and more obstacles.}
\label{test_env}
\end{figure*}

The objective of the robot is to move in turn to the red destinations shown in figure \ref{test_env}. The evaluation metrics for these scenarios include the number of destination reached, final return, traveling time, and traveling distance. We performed each method, which is either baseline or proposal method training in either Env-1 or Env-2, five times in each scenario, and took the average results. In addition, if the robot can reach all the destinations, other metrics will not be mentioned. Table \ref{test1_res} and \ref{test2_res} depicts robot's performance in evaluation scenario 1 and 2, respectively.\newline
\begin{table}[h!]
\begin{center}
    \caption{Robot's performance in evaluation scenario 1}
    \begin{tabular}{|C{1.2cm}|C{1.2cm}|C{1.2cm}|C{1.2cm}|C{1.2cm}|}
        \hline
        \textbf{Method} & \textbf{Finished Destinations}  & \textbf{Final return}  & \textbf{Traveling time (s)} & \textbf{Traveling distance (m)} \\
        \hline
        Baseline + Env-1 & 2& - & - & -\\
        \hline
        \textbf{Proposal + Env-1}& 1&  - & - & -\\
        \hline
        \hline
        Baseline + Env-2& 6 & 7621.67 & 132 & 17.50\\
        \hline
        \textbf{Proposal + Env-2} & \textbf{6}&  \textbf{7538.65} & \textbf{106} & \textbf{17.08}\\
        \hline
    \end{tabular}
    \label{test1_res}
\end{center}
\end{table}

\begin{table}[h!]
\begin{center}
    \caption{Robot's performance in evaluation scenario 2}
    \begin{tabular}{|C{1.2cm}|C{1.2cm}|C{1.2cm}|C{1.2cm}|C{1.2cm}|}
        \hline
        \textbf{Method} & \textbf{Finished Destinations}  & \textbf{Final return}  & \textbf{Traveling time (s)} & \textbf{Traveling distance (m)} \\
        \hline
        Baseline + Env-1 & 1& - & - & -\\
        \hline
        \textbf{Proposal + Env-1}& 1&  - & - & -\\
        \hline
        \hline
        Baseline + Env-2& 6 & 10548.76 & 194 & 24.05\\
        \hline
        \textbf{Proposal + Env-2} & \textbf{6}&  \textbf{10591.78} & \textbf{157} & \textbf{25.99}\\
        \hline
    \end{tabular}
    \label{test2_res}
\end{center}
\end{table}

It can be seen that the final returns of the proposed method are around these of baseline  (7621.67 and 10548.76 compared with 7538.65 and 10591.78 of the baseline), while the traveling time and the traveling distance is even lower (157 (s) compared with 194 (s) of the baseline, in scenario 2, and 17.08 (m) compared with 17.05 (m) of the baseline, in scenario 1). However, this also does not prove that the proposed method is any better. The notion that we want to pinpoint here that our approach can well approximate the baseline in performance with low-cost hardware. The results demonstrated in table \ref{test1_res} and table \ref{test2_res} also indicate the stability of our proposed method.

\section{Conclusion and future works}
This work implements a mapless planner for the mobile robot with a state-of-the-art deep reinforcement learning algorithm. We also propose to use images from a monocular camera, replacing the laser findings, for the purpose of saving costs for the hardware equipment. The experimental results show that the proposal's performance is as satisfactory as the baseline is.\newline

Deep reinforcement learning is flourishing as recent works have been coming up with novel ideas and strategies for solving key problems in reinforcement learning such as sampling efficiency or exploration-exploitation trade-off. These algorithms are often applied to video game environments as well as other physics emulators but not much to robot navigation and obstacle avoidance. Accordingly, we will work on these ones to solve our problem and hope to achieve more impressive performance in not only enhancing robot's behavior but also reducing searching space.
\section*{Acknowledgment}
This research is supported by research funding from Honors Program, University of Science, Vietnam National University - Ho Chi Minh City.

\bibliographystyle{unsrt}
\bibliography{references.bib}

\end{document}